\documentclass[nonacm]{acmart}

\usepackage{algorithm}
\usepackage{algpseudocode}
\usepackage{natbib}
\setcitestyle{square, comma, numbers,sort&compress, super}

\begin{document}

\title{A Customizable Generator for Comic-Style Visual Narrative}

\author{Yi-Chun Chen}
\affiliation{%
  \institution{Computer Science, North Carolina State University}
  \country{USA}
  }
\email{ychen74@ncsu.edu}

\author{Arnav Jhala}
\affiliation{%
  \institution{Computer Science, North Carolina State University}
  \country{USA}
  }
\email{ahjhala@ncsu.edu}







\renewcommand{\shortauthors}{Chen et al.}

\begin{abstract}
We present a theory-inspired visual narrative generator that incorporates comic-authoring idioms, which transfers the conceptual principles of comics into system layers that integrate the theories to create comic content. The generator creates comics through sequential decision-making across layers from panel composition, object positions, panel transitions, and narrative elements. Each layer's decisions are based on narrative goals and follow the respective layer idioms of the medium. Cohn's narrative grammar provides the overall story arc. Photographic compositions inspired rule of thirds is used to provide panel compositions. McCloud's proposed panel transitions based on focus shifts between scene, character, and temporal changes are encoded in the transition layer. Finally, common overlay symbols (such as the exclamation) are added based on analyzing action verbs using an action-verb ontology. We demonstrate the variety of generated comics through various settings with example outputs. The generator and associated modules could be a useful system for visual narrative authoring and for further research into computational models of visual narrative understanding.
\end{abstract}

\maketitle

\section{Introduction}

As a popular visual medium, comics are a dominant form of storytelling across cultures and age groups. They integrate textual expressions with graphical representations to communicate intricate stories through multi-modal panel sequences. Comic book authors play around with visual elements for effective storytelling\cite{martens2020visual}. Interactive visual novels are also a popular genre among interactive storytelling media. In this paper, we introduce a customizable end-to-end digital comic generator. Our generator organizes the process of generating comics by providing systematic components that incorporate narrative arcs, character and scene parameters, and page and panel composition idioms. This knowledge is accessible through an API for plugging in different rendering engines. The automated generator reasons about narrative through a narrative grammar module that encodes narrative arc proposed by Cohn\cite{cohn2014grammar}. Consistent with the placement of panels based on the grammar, selection of objects in terms of inclusion/exclusion of panels is done followed by their placement on the panels based on the rule-of-thirds composition rules. Using a database of verbs and their sentiment extracted from the circumplex model, the action layer scores compositions based on emotional appeal for individual panels. This is followed by an evaluation of the entire sequence's score on the narrative arc. Panel placement is based on the number of panels and type of transitions between panels proposed by McCloud\cite{mccloud1993understanding}.

Generating comics in a modular generator that encodes layer-wise theories helps demonstrate the effectiveness of the underlying methods and also opens up possibilities of extending and comparing different representations for their expressiveness. This also provides a source of generating many comics that have similar visual content for studying comprehension and panel composition with controlled content\cite{swanson2012learning}. 


The comic generator is modular with an application program interface (API) support for setting and modifying generator parameters by layer. Modifications on either comic sequence, panels, or intra-panel content can be applied by adding customized refinement layers. To start off, we provided a small dataset of visual materials that support comic-style image content based on the familiar image representations or common abstractions in real comics. The sample layers show the possibility of combining rules and visual elements to comic content.

This work contributes to the community in three ways. First, to provide a platform for encoding and testing visual narrative theories and storytelling through generative methods. Second, to organize the generation process of complex visual narratives into well-defined layers with clear semantics (not formal logical but human understandable) that map to a human comic author's workflow. Third, we provide an initial domain and encoding of current theories for demonstration.

\section{Related work}

Narrative generation research and practice often separates out story plot specification from realization or telling\cite{chatman1980story}\cite{jhala2005discourse}. This allows researchers to separate out reasoning about plot structure and apply constraints on media selection to support the authorial goals with respect to the generated plots. Comics contain both visual media components and language components including stylized annotations and panel compositions. Comic artists have documented their process and principles from their experience. Eisner's book discusses the diversity of comics' image representations \cite{eisner2008comics}; Carrier et al.\cite{carrier2001aesthetics} provides a commentary on the aesthetics of comic strips as a form of fine art. Scott McCloud's work, which provides a comprehensive treatment of the many aspects of comic development, such as panel transitions that capture the focus changes between consecutive panels, the expression of action over time, and the symbolic metaphors that make abstract ideas visible \cite{mccloud1993understanding,mccloud1998understanding,mccloud2006making}. Besides the principles that apply to graphical aspects, the underlying structure of visual narrative has also received focus recently. Cohn et al.\cite{cohn2013visual,cohn2014grammar,cohn2015analyze,cohn2016visual} provide a comprehensive treatment of the cognitive processes from low-level perception through high-level discourse and organization\cite{freytag1908freytag,stern2000making,pratt2009narrative} with their proposed theory of the Visual Narrative Grammar (VNG).

Besides discussing comic structures and visual grammar,  other researchers have developed models or systems to generate sequential art that incorporates different features based on the analyzed theories and ideas. Alves et al.\cite{alvesgenerating} discussed the graphical composition of panels by describing the camera placement, background, independence, and other features. In comparison, Nairat et al.\cite{nairat2020generative} focus more on the fictional part of comic. They propose inner character evolutions to form stories which are then rendered as comics. Martens et al.\cite{martens2016discourse,martens2016generating} propose a visual narrative engine from the point of view of generative systems for the discourse of comic sequences. Their sample generator uses abstract shapes for demonstrating the variety of stories that can be communicated.

Our work aims to provide a generator that integrates and tests multiple aspects of comics rather than focusing on only one element to tackle the challenge result from the complexity of this visual medium. Our generator divides the complex generation problem by systematically organizing layers of decision making for composing comics. To achieve this, the generator supports the functions that can modify elements obtained from decomposing either comic's graphical composition and the structure as well as discourse elements, which makes the generator extendable and modular. We demonstrate how to encode concepts from a sample set of theories into the generator and provide sample dataset curated by retrieving common visual metaphors from real comic datasets such as COMICS \cite{iyyer2017amazing}\cite{chen2021computational} and Manga109\cite{narita2017sketch,matsui2017sketch}.

\section{Method}



\begin{figure*}[ht]
    \centering
    \includegraphics[width=0.6\linewidth]{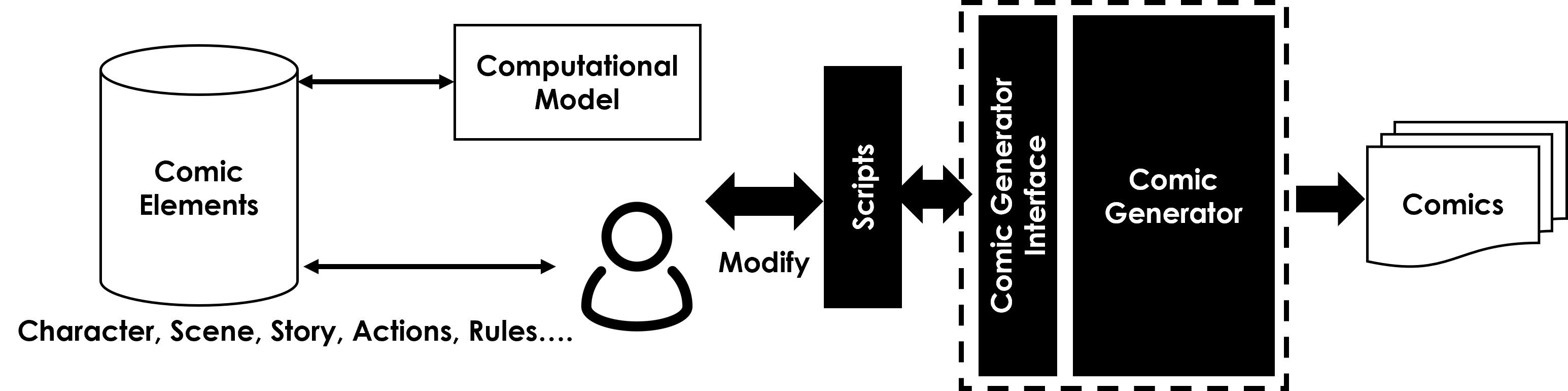}
    \caption{Overall software architecture of the generator}
    \label{big_picture_2}
\end{figure*}

The generator is structured to follow the human workflow of generating a comic (Figure \ref{big_picture_2}). The inputs to the generator are a script that includes action descriptions of characters and the elements information such as graphic composition, rules, and reference values. The generator renders panels on pages by querying the authoring idioms via an application programming interface (API). The composition process starts with the determination of the number of panels that are then overlayed on an encoding of the high-level narrative arc with labels proposed by Cohn (details are in section 4). The next layer focuses on the composition of individual panels based on the number of included objects composed in terms of size and position by following the rule of thirds from photography. Once panel compositions are chosen, the transition layer determines the relationship between adjacent panels according to the four types of transitions described by McCloud. Finally, the last layer uses action verbs to choose overlays and emotion symbols. Once the panel composition and sequence are complete, the overall sequence is provided a score that is used to visualize the narrative arc.


\subsection{Overall Structure of the Generator}



\begin{figure*}[ht]
    \centering
    \includegraphics[width=0.8\linewidth]{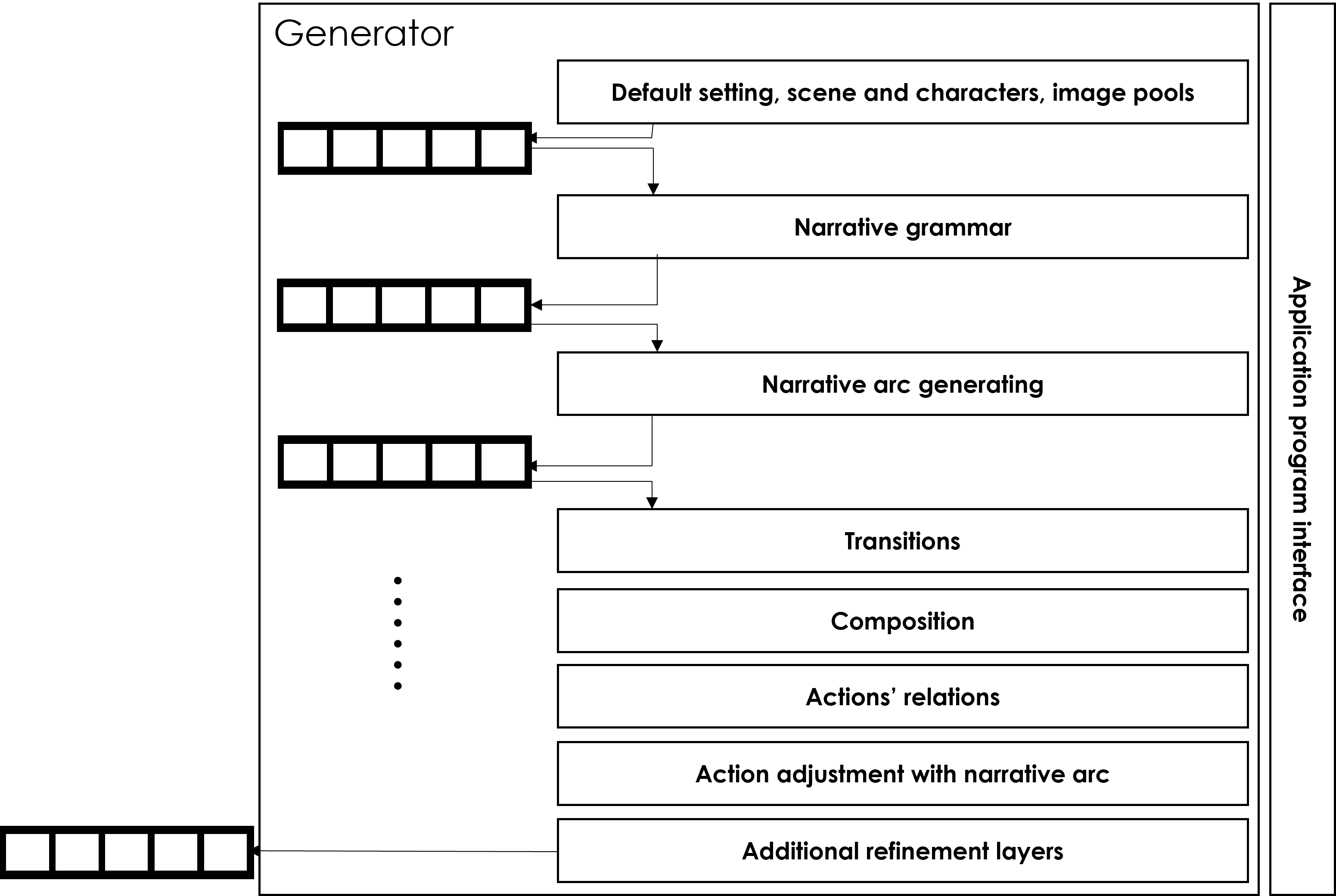}
    \caption{Sequence of operations in the generator}
    \label{Generator_structure}
\end{figure*}

Our generator modifies the comic sequence based on iteratively applying refinement layers. It first creates an empty comic sequence with a random number of panels. Afterward, according to the settings or rules in the refinement layers, the model keeps modifying the content. The modifications start with the whole sequence to fit it to a narrative structure. In addition to the set of panels and panel sequence, additional refinement layers can enlarge the generating base by adding new elements to enrich the comic content or add more diversity to the image representations. For example, a layer applies the mapping between textbox shapes with character actions' activation value. A layer provides different image composition templates. The generator provides a set of application program interfaces (API) to allow extendability. The generator's structure and an example of combined refinement layers through API are in figure \ref{Generator_structure}. It shows the combination of sample layers we describe in this paper and shows a sample script.


\subsubsection{API Design}

\begin{figure*}[ht]
    \centering
    \includegraphics[width=\linewidth]{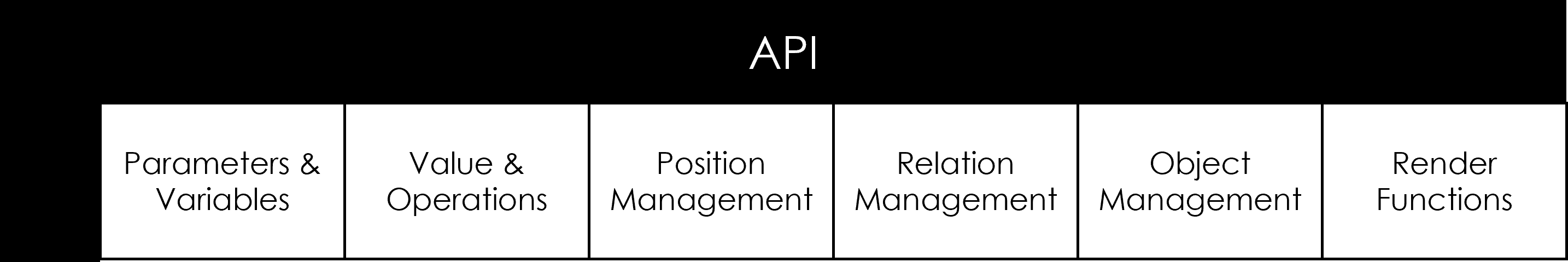}
    \caption{Sections of the API}
    \label{API_and_functions}
\end{figure*}

Figure \ref{API_and_functions} provides the view of  functionality modules inside the generator. The functionalities of the generator can be divided into several main chucks: object management, relation management, position management, basic operations on objects and relations, and the parameters inside the generator.

\begin{itemize}
    \item \textbf{Object Management:} This part provides functions that manage the object's pool, the elements that be used in comics, that is used in comics such as possible actions, graphical materials, and panel compositions.
    \item \textbf{Relation Management:} It controls the relations between objects. for example, the causal links between select-able actions. 
    \item \textbf{Position Management:} This module divides comic panels into coordinates with two dimensions, and this is the position reference for rendering the graphical elements.
    \item \textbf{Render Function:} It renders graphical elements for comic panels according to layer and position settings. The background, foreground, and the entities of comics such as symbols and characters are separated based on the information.
    \item \textbf{Value and Operations:} This part provides simple operations such as subtraction and addition for operating on assigned values for objects.
    \item \textbf{Parameters:} The settings and variables of the comic generator.
\end{itemize}

\subsection{Generating process: Refinement Layers with Show Cases}
    This document presents several sample layers of the generator to show its ability to combine with customized rules. We briefly introduce each layer below, but more details will be described further in the following sections. Here are how the refinement layer serves on generating comic. This section includes showing cases that how a comic sequence will look after modifying by new refinement layers and also introduce how the functions inside the module are applied to layer settings. 

\begin{figure*}[ht]
    \centering
    \includegraphics[width=\linewidth]{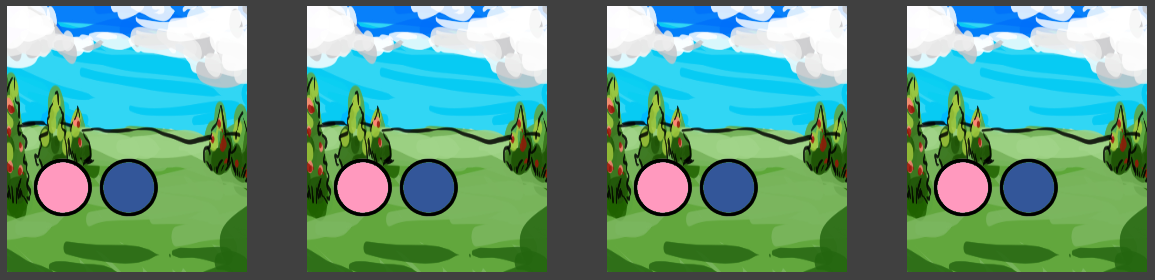}
    \caption{A comic sequence with default randomized settings.}
    \label{default_sequence}
\end{figure*}

    In the beginning, if no refinement layer is added, the generator will generate a panel sequence with a random length according to the basic settings: two characters, a scene, and a default composition. Then each refinement layer gradually modifies the content. \ref{default_sequence}

\begin{itemize}

    \item \textbf{Grammar Layer } The first layer integrates Cohn's Visual Narrative grammar in our demo settings. It provides several category labels and proposes a tree structure for future comic content. And it applies the functions in the parameter \& variables module to record the category labels. Then it implements a tree expansion function to create a center-embedded tree as a reference for the comic sequence.
    \item \textbf{Narrative Arc Generating Layer:} The layer links values with the grammar layer. It calls functions in parameter \& variables module to retrieve VGN phases and assign tension scores to it as a reference for comic content. 
\end{itemize}     

    These two layers display the ability to add references for creating the content, which builds the basis for making decisions in afterward layers.

\begin{itemize}    
    \item \textbf{Action Layer:} The next layer is an Action Layer; it applies the functions in the Object Management module and the Relation Management module to create a small causal relation network among sample actions. Then, each panel implements a selection between actions according to the small action network. We then have two variants of the action layer: one selects actions only based on the given action network, and the other choosing actions refers to the reference that the narrative arc layer provided. In the latter layer, the value \& operations module functions are used to bring possible tension value with each action verb. Finally, each action links to symbols that we provide according to the common symbol used in real comics.
\end{itemize} 

    The apply of action network, give the generator a basis for content selection with some causal relation among the action pool\ref{action_net_layer}.The version with narrative arcs shows in following sections.
    \begin{figure*}[ht]
        \centering
        \includegraphics[width=\linewidth]{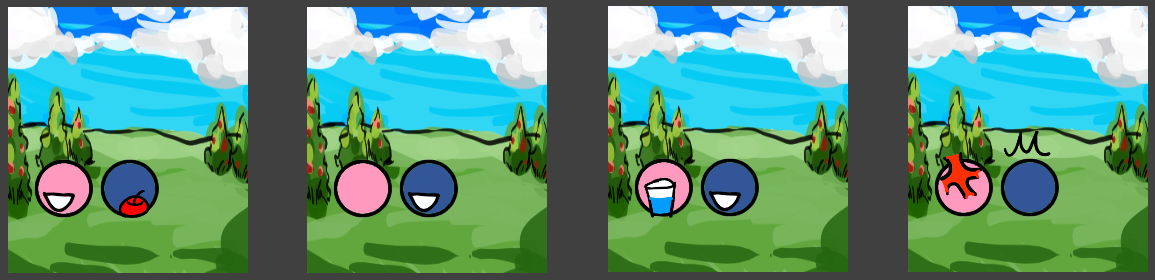}
        \caption{A comic sequence with actions.}
        \label{action_net_layer}
    \end{figure*} 

\begin{itemize}
    \item \textbf{Composition Layer:}To add more changes to the graphical representation of comic panels, we then customized a layer that reads the position data through functions in the position management module. The functions help this refinement layer transform the position into different panel layouts. The panel layouts that are imported from the composition layer bring variation of how the characters combines the scene together which adds more possibility to content.
    
    \item \textbf{Transition Layer:} This layer integrates the composition information from the Composition Layer and setting conditions to change the action selections further. It follows the conceptual focus changing types( McCloud's panel transitions) which will be described in detail in other sections. 
\end{itemize}
    With the cooperation of the two layers, the generator now has the ability to modify the focus between panels with a more flexible panel composition.

\begin{figure*}[ht]
    \centering
    \includegraphics[width=\linewidth]{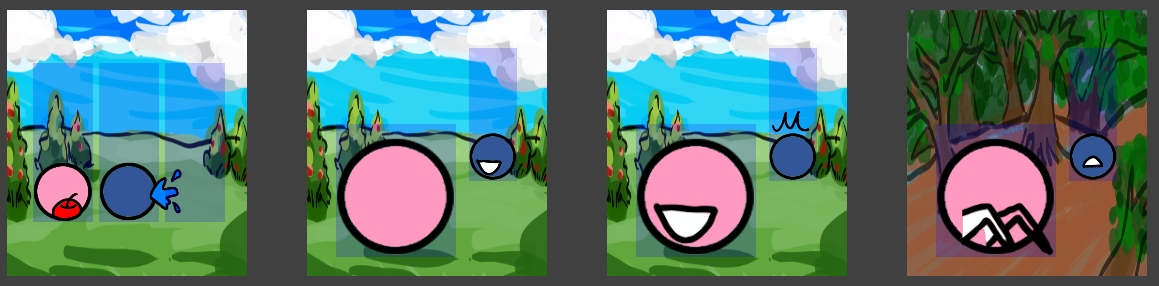}
    \caption{A comic sequence with composition information and panel transitions.}
    \label{character_with_layout}
\end{figure*}

 \ref{character_with_layout}
    
\begin{itemize}    
    \item \textbf{Customized Layer:} This sample layer gives cases that more graphical symbols or customization of graphical representation can change the comic sequence. It links a new symbol pool of various types of textboxes, so the results can show up on the new comic sequence.  
    \item \textbf{Story Layer:} This layer combines the Object Management module and the Relation Management module to take story statements, the assertions in Harman's Rensa system \cite{harmon2017narrative}. The story texts, after being processed into predicates forms, can be imported into the generator. By creating links between entities and actions inside the story. The story layer equips the ability to modify the content according to the story text \{A and B are in the forest. A has B's Apple. B is angry. B enter the scene.\}.
\end{itemize}
    
    The customized layer add some more diversity to the graphical representations, and the Story layer provides further ability to the generator that it can now generate comic sequence according to certain content. This change makes the comic sequence show a rather complete story. 
    
\begin{figure*}[ht]
    \centering
    \includegraphics[width=\linewidth]{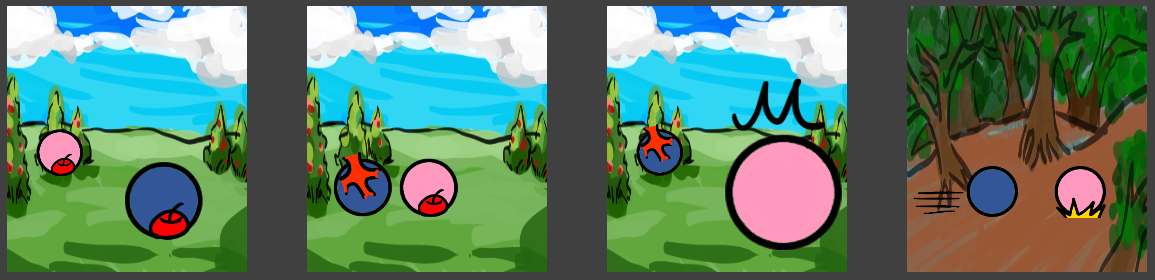}
    \caption{A comic sequence with story content}
    \label{comic_with_story}
\end{figure*} 

\subsection{Graphical Content}

We observed how existing comics represent specific actions and emotions with symbols to generate the image representation. Then combine the abstracted symbols of actions with character representation and scenes \cite{alvesgenerating}. 



\subsubsection{Scene}


The scene of a comic panel is a particular place where an event happens. In our sample materials, five representative scenes are chosen as examples to support the generating process.
\subsubsection{Abstractions}
Our model used abstract symbols to present the story's characters. Similar to Martens et al.\cite{martens2016generating} where they used geometric shapes to represent characters in their comic generating work, we also employ simple shapes to communicate the complex emotional details of characters.
\subsubsection{Common Symbols}


In comics, authors usually employ several abstracted symbols to exaggerate characters' emotions and actions, such as speed lines to show the movement, explosion-shape to represent objects' collision, cross-shape symbols to emphasize anger, etc. Table \ref{symbol} shows some examples from real comics. We analyzed the representations of actions and emotions from real comics. Those elements provide visible hints to readers about character status and changes. 

\begin{table}
\centering
\begin{tabular}{|p{1.4cm}|p{1.4cm}|p{1.4cm}|p{1.4cm}|p{1.4cm}|p{1.4cm}|p{1.4cm}|p{1.4cm}|}
    \hline
    Anger & Quick moving & Slow moving & Anxious & Collision & Relieved & Shock & Big shock\\
    \hline
    \includegraphics[width = \linewidth]{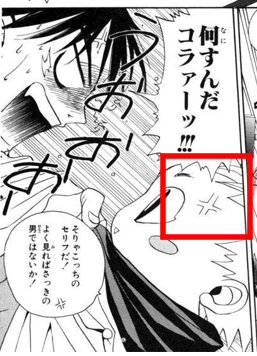}
    &\includegraphics[width = \linewidth]{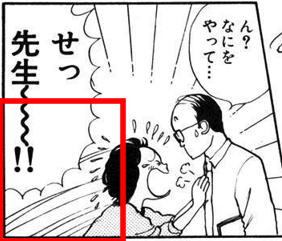}
    &\includegraphics[width = \linewidth]{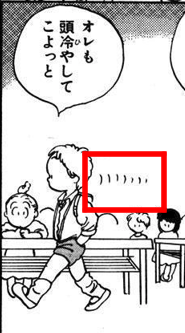}
    &\includegraphics[width = \linewidth]{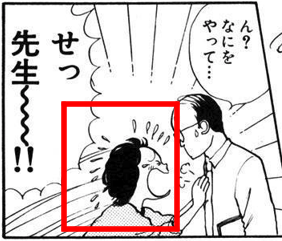}
    &\includegraphics[width = \linewidth]{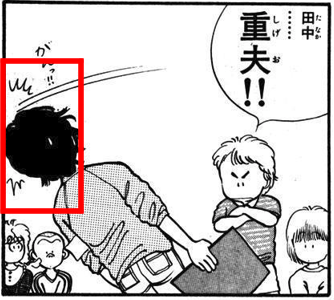}
    &\includegraphics[width = \linewidth]{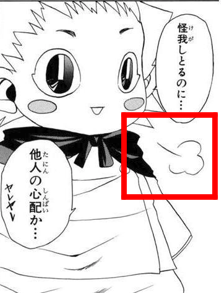}
    &\includegraphics[width = \linewidth]{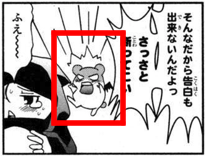}
    &\includegraphics[width = \linewidth]{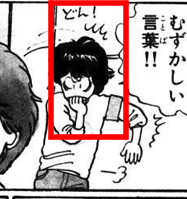}\\
    \hline     
\end{tabular}
    \caption{
    \label{symbol} \small{Examples of metaphor symbols of action and characters' emotions from real comics.AisazuNihaIrarenai© Yoshimasako, AkkeraKanjinchou© Kobayashiyuki,Akuhamu© Araisatoshi}
    }
\end{table}

\section{Generator Usage: Creating Layers}

This section provides examples of layers to show how to use our model and API to generate comics. The sample layers show three scopes--sequence,  panel, and intra-panel content--of modification that the model can perform and how those combine with comic theories, hence applying the principles to comics.

\subsection{Sequence Modifications}
To generate a comic sequence, what needs to be considered are not only the visual representations in panels but also the relation between consecutive content. Therefore, we combined the narrative grammar that describes the overall structure of comic sequences with the concept of narrative arc or emotional trajectory. Each panel's composition is constrained by the included content and how it is composed with respect to the emotional trajectory of the story progression.

\subsubsection{Narrative Grammar}
We formalize the generation process of new comics globally at the story level and locally at the panel level. This maps to factors that influence either the whole sequence or only individual panel content. Starting from the comic sequence's structure to decide the content's global reasoning, we generate the narrative structure of the target. We adopt Cohn's narrative structure \cite{cohn2013visual} that proposed that coherent comics follow a grammar that organizes its global structure with five categories: 

\begin{itemize}
	\item \textbf{Establisher (E)} - The settings of characters, place, etc., without action involved.  
	\item \textbf{Initial (I)} - It is the beginning of a story arc--the start of an action or an event.
	\item \textbf{Prolongation (L)} - The Middle state of the story arc--extends an action.
	\item \textbf{Peak (P)} - The highest story tension--the end of an action.
	\item \textbf{Release (R)} - It releases the tension—the consequence of an action.
\end{itemize}
And the five categories form basic phases through linear ordering:

	Phase (Establisher) - Initial(Prolongation) - Peak - (Release)

The parentheses above imply these categories are optional. The importance of each type from high to low is Peak, Initial, Release, Establish, and Prolongation. Furthermore, more complex combinatorial structures can be constructed through conjunctions of embeddings. Our model generates the narrative structure through center-embedding; it expands a new tree structure by replacing a single category with a phase.  

We start with a primary phase (ex. E, I, P, R) and form a tree structure according to the chosen phase where the categories are leaf nodes. And then, referencing the center-embedded pattern of VGN theory, the model expands nodes with phases depending on probability. Thus, we can get a structure tree whose leaf nodes represent panels of a comic sequence. Table \ref{grammar} gives an example of comic sequences following the grammar structure.

\begin{table}
\centering
\begin{tabular}{|p{1.5cm}|p{1.5cm}|p{1.5cm}|p{1.5cm}|p{1.5cm}|}
    \hline
    E & I& L& P & R\\
    \hline
    \includegraphics[width = \linewidth]{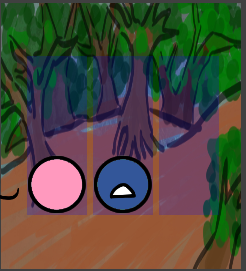}
    &\includegraphics[width = \linewidth]{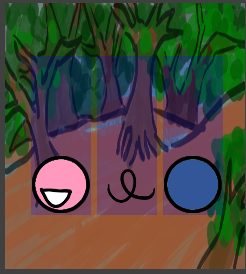}
    &\includegraphics[width = \linewidth]{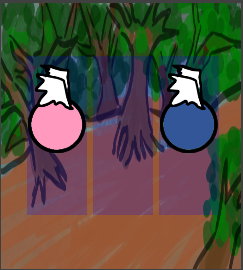}
    &\includegraphics[width = \linewidth]{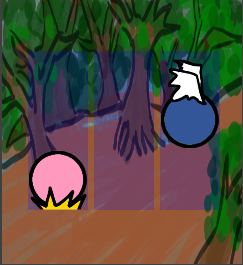}
    &\includegraphics[width = \linewidth]{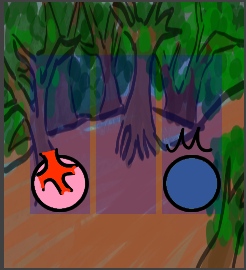}\\
    \hline     
\end{tabular}
    \caption{
    \label{grammar} \small{Examples of a comic sequence that followed narrative grammar.}
    }
\end{table}

\textbf{Narrative Grammar Layer Customization}

Here is the flow that how we decide narrative structure \ref{grammar_algo}. The setting of the grammar layer is to decide the narrative structure of the comic sequence. Therefore, after creating an object dictionary for grammar phases and expending the tree structure for the sequence, the sequence length is adjusted according to the structure. In the layer, it assigns a grammar phase to a panel to indicate the narrative function it should perform in these sequences which become the references when selecting actions as content.  

\begin{algorithm}
\caption{Grammar Layer}\label{grammar_algo}
\begin{algorithmic}[1]
\Procedure{Grammer Layer Script}{}
\State $P$ $\gets$ input the current panel sequence.
\State $VGN$ $\gets$ Create object list for VGN basic phases.
\State $S$ $\gets$ Expend a center-embedded tree with $VGN$, get the reference narrative structure.
\If Length($P$) $\neq$ Length($S$)
    \State Add or Subtract empty panels from $P$  
\EndIf
\State \emph{loop:}
    Assign each phases in $VGN$ to $P$
\State \Return $P$

\EndProcedure
\end{algorithmic}
\end{algorithm}

\subsubsection{Narrative Arc}
The five categories in narrative grammar reflect the concept of narrative arc, which describes a story's full progression. It implies that every story has a relatively calm beginning and then reaches the highest tension in the middle where character conflict and narrative momentum happens; In the end, conflict is resolved. This paper projected the five grammar categories to a value to get a curve that describes how the story tension changes.

\textbf{Narrative Arc Layer Customization}
The process of how the narrative arc combine with grammar structure is listed in \ref{narrative_arc}. Because the visual grammar phases indicate the narrative function of each panel in a sequence, the structure becomes a reference for the tension changes of the generated story--the narrative arc. To capture the abstract concept of tension we mapped each grammar phase with different scores between one to ten, according to the narrative functions. For example, the Peak (P) should be the one with the highest tension and the Release (R) loosen the emotion to some extent. Therefore, we start the setting of this layer from create a value dictionary for each grammar phase. Then the next step is to map the grammar phase the sequence got from previous layers into the score and create the curve of the narrative arc.  

\begin{algorithm}
\caption{Narrative Arc}\label{narrative_arc}
\begin{algorithmic}[1]
\Procedure{Mapping Narrative Arc with Structure}{}
\State $P$ $\gets$ input the current panel sequence.
\State $REF$ $\gets$ create value dictionary with VGN phases with assigned tension scores.
\State $REF = \{ E:0 I: 2 L: 4 P: 6 R:2\}$
\State $OPER = \{ E:EQUL I: ADD L: ADD P: ADD R: Equal\}$
\If {panels in $P$ have assigned grammar phase}
\State \emph{loop:} over $P$, apply $REF$ with $OPER$ with the grammar phase 
\Else {use default narrative arc scores}
\EndIf
\State \Return $P$

\EndProcedure
\end{algorithmic}
\end{algorithm}


\subsection{Panel Modifications}
This section discusses the modifications that apply to the whole panel: panel compositions and the panel transitions.
\subsubsection{Compositions of Panels}
We applied the composition concept in photography to comic panels to simplify the diversity of panel content composition. Instead of using continuous coordinates to describe positions, we divided our panel into nine parts according to the elemental composition method in photography, which is the rule of third. Based on the four intersection points, we created two masks as the abstractions of character positions. Each mask was divided into three height levels; these are used to present the vertical physical position changes. The horizontal position changes were the shifts between masks set.  

We used the arrangement of masks to achieve simple view angle changes and composition. Our model had two versions: the basic version provided the physical changes due to action changes; the other version then combined the view change to create more tension in the comic panels. Table \ref{composition} shows the masks for image compositions we used in our model.
\begin{table}
\centering
\begin{tabular}{|p{1.4cm}|p{1.4cm}|p{1.4cm}|p{1.4cm}|p{1.4cm}|p{1.4cm}|p{1.4cm}|}
    \hline
    Rule of third & Basic & Parallel view& Left view with medium shot& Right view with medium shot & Left view with close shot & Right view with close shot\\
    \hline
    \includegraphics[width = \linewidth]{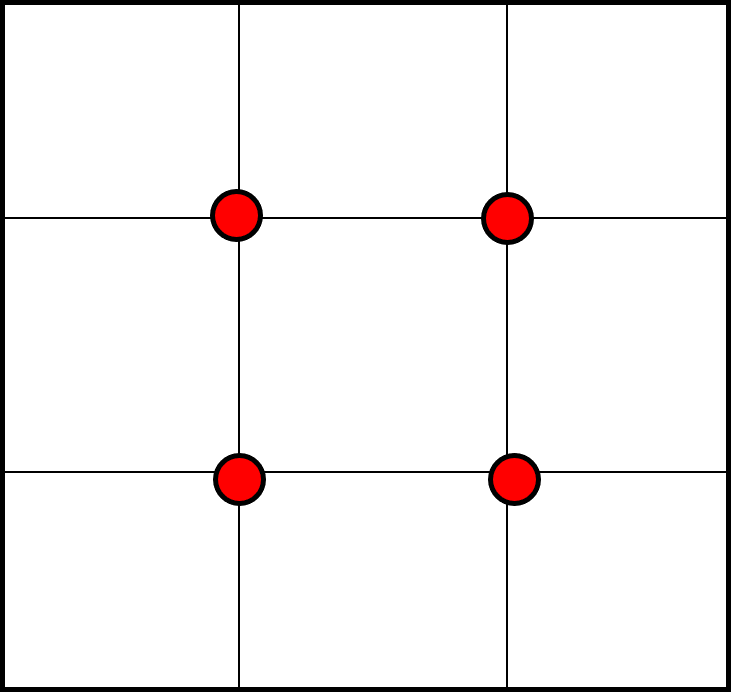}
    &\includegraphics[width = \linewidth]{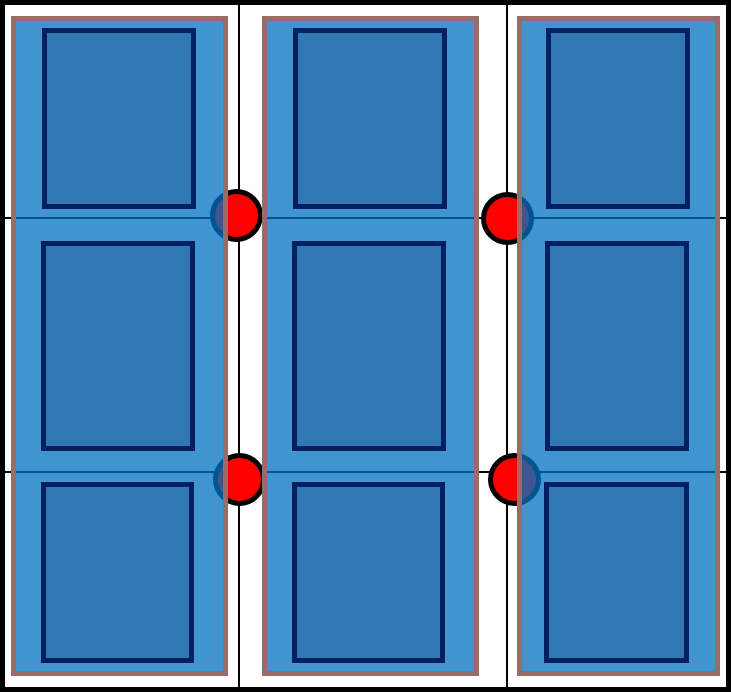}
    &\includegraphics[width = \linewidth]{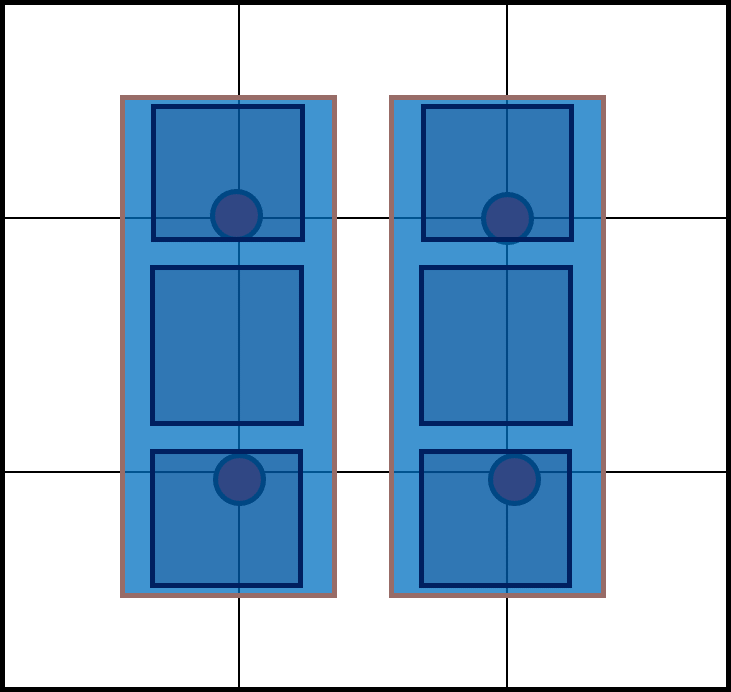}
    &\includegraphics[width = \linewidth]{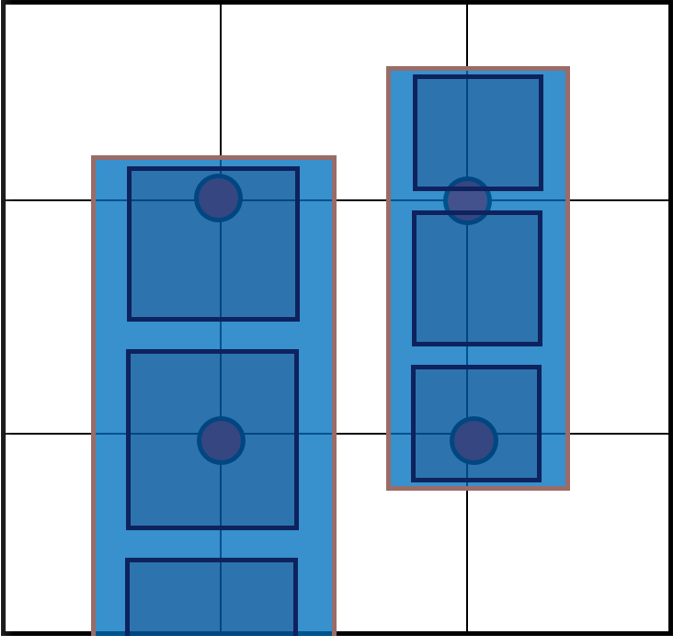}
    &\includegraphics[width = \linewidth]{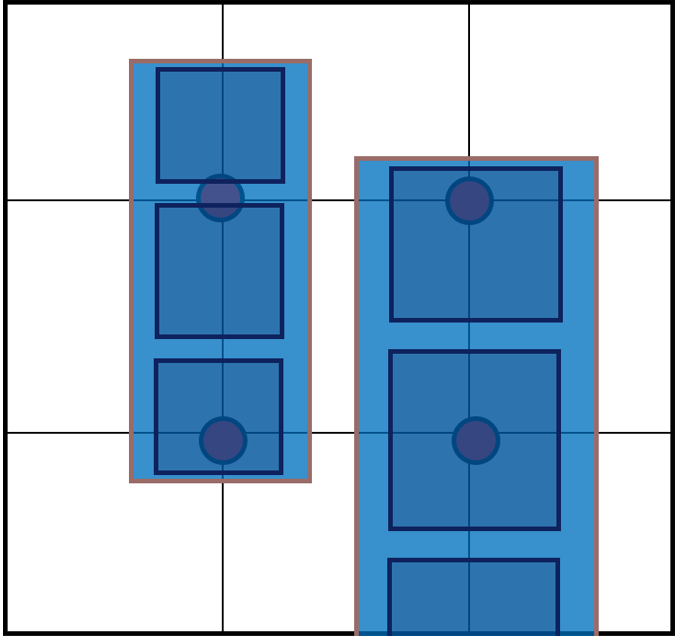}
    &\includegraphics[width = \linewidth]{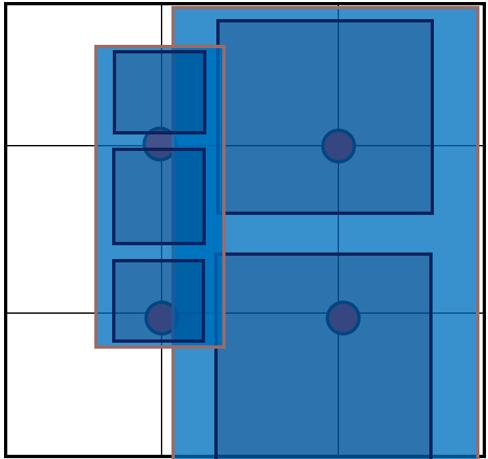}
    &\includegraphics[width = \linewidth]{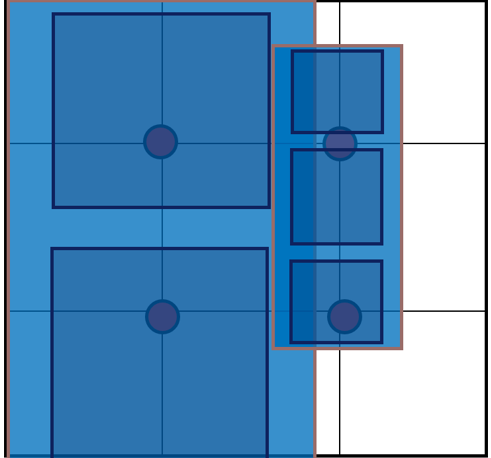}\\
    \hline     
\end{tabular}
    \caption{
    \label{composition} \small{Examples of geometric abstractions and image compositions.}
    }
\end{table}
\textbf{Layer Settings}
\subsubsection{Transitions}
In McCloud's comic theory \cite{mccloud1993understanding}, the transitions that bridge comic panels and describe the relation of gutters can be placed in several categories. These transitions are how the focus shifts in consecutive panels. We employed the following transitions to plan the shift between image content. 
\begin{itemize}
	\item \textbf{Action-to-Action} - This transition featuring a single subject in distinct action progression. We applied this to choose action for next panel.
	\item \textbf{Moment-to-Moment} - This transition captures slight changes in time and space. We use it to restrict action selections.
	\item \textbf{Aspect-to-Aspect} - This transition sets a wandering eye on different aspects of a place, or idea. When applying, we link it with image compositions. 
	\item \textbf{Scene-to-Scene} - This transition transport readers across a significant distance of space. Our model uses it to rearrange the scene.

\end{itemize}
Table \ref{transition} shows an example of how the image content changes after applying each transition correspondingly. 
\begin{table}
\centering
\begin{tabular}{|p{1.4cm}|p{1.4cm}|p{1.4cm}|p{1.4cm}|p{1.4cm}|p{1.4cm}|p{1.4cm}|p{1.4cm}|p{1.4cm}|}
    \hline
     Action Pre & Action After&  Aspect Pre & Aspect after& Scene Pre & Scene after & Moment Pre & Moment After\\
    \hline
    \includegraphics[width = \linewidth]{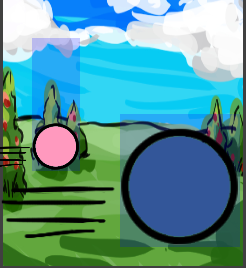}
    &\includegraphics[width = \linewidth]{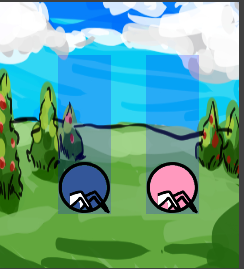}
    &\includegraphics[width = \linewidth]{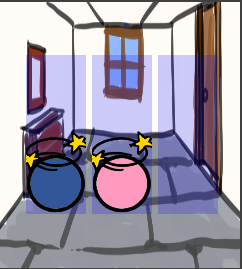}
    &\includegraphics[width = \linewidth]{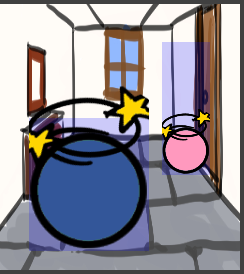}
    &\includegraphics[width = \linewidth]{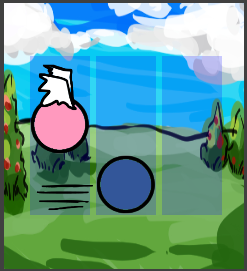}
    &\includegraphics[width = \linewidth]{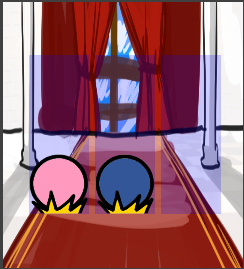}
    &\includegraphics[width = \linewidth]{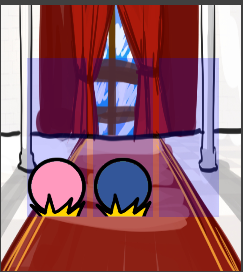}
    &\includegraphics[width = \linewidth]{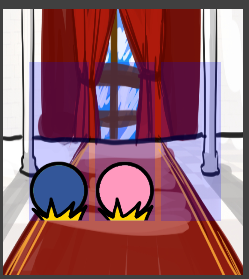}\\

    \hline     
\end{tabular}
    \caption{
    \label{transition} \small{Examples of applying transitions between panels.}
    }
\end{table}

\textbf{Composition Layer Customization}
We link the panel transitions to panel compositions in this layer \ref{transition_layer_algo} because panel transitions capture the focus change among panels, whereas films usually use different shots to achieve this. This is why the setting starts by creating the links between various transitions and composition changes. After randomizing the transition sequence, the layer makes selections among possible image compositions after applying the transition.

\begin{algorithm}
\caption{Transition}\label{transition_layer_algo}
\begin{algorithmic}[1]
\Procedure{Generate Transition Sequence}{}
\State $P$ $\gets$ input the current panel sequence.
\State $TRANS$ $\gets$ create object dictionary for McCloud's transitions.
\State $COMP$ $\gets$ create object dictionary for sample panel compositions.
\State $REF$ $\gets$ create relations between $TRANS$ and $COMP$.
\State \emph{loop:} over $P$, random design a transition sequence 
\State \emph{loop:} random select a composition according to $REF$ 
\State \Return $P$
\EndProcedure
\end{algorithmic}
\end{algorithm}

\subsection{Intra-panel Modifications}
Besides the structure that influences the whole sequence and elements that change panels, the semantics in each panel are also important. This section will describe how we plan the intra-panel changes, in other words, content inside a panel with the help of the character's actions and action symbols.

\subsubsection{Action Network}

We recorded common metaphor symbols from real comics and observed nineteen actions and reactions. And then, we defined an action relation net based on the actions, forming a graph representing possible causal relations between characters' actions. Table \ref{actions} lists part of the basic actions we have in our model and their reactions. The relations between actions and their reactions predicate the possible outcomes of the actions; hence the model can decide what will possibly happen in the next panel. For example, if the character is sitting, it will not start running immediately and will not start rolling or falling too. Similarly, if a character is falling, it will not fall asleep right after falling. Actions and their possible result form the action relation net in our model. It can be modified and extended by providing new actions and links.
\begin{table}
\centering
\begin{tabular}{|p{1.8cm}|p{10cm}|}
    \hline
    Actions & Reactions\\
    \hline
    Stand& [Stand], [Sit], [Upset], [Laugh], [Mad], [Shock], [Walk], [Run], [Jump], [Dizzy], [Worry], [Think],  [Relief]\\
    \hline
    Fall& [Upset], [Mad], [Shock], [Collide], [Dizzy], [Worry]\\
    \hline     
    Jump& [Stand], [Fall], [Collide]\\
    \hline     
    ...& ...\\
    \hline  
\end{tabular}
    \caption{
    \label{actions} \small{Part of sample actions and their reactions.}
    }
\end{table}

When constructing the content of a new panel, another essential thing is what the characters are doing is also important. Besides the characters and the scene, the characters' actions decided the semantics of the panel. More accurately, what the characters perform in the panel--the actions form the whole story content and make the sequence reasonable. Therefore, what happens in each panel was planned according to the action relation net. The action in the last panel decides the possible candidate of action in this panel.

\textbf{Action Network Layer Customization}
This layer links selectable actions with causal relations and applies the links into content \ref{action_net_algo}. It imports the causal relation network between actions. For example, a dizzy action may show up after a collide action; or only after performing a jump action, a character can fly. The layer refines the panel content according to the causal network. Each action is chosen according to the previous actions.

\begin{algorithm}
\caption{Action Network}\label{action_net_algo}
\begin{algorithmic}[1]
\Procedure{Generate Action Network}{}
\State $P$ $\gets$ input the current panel sequence.
\State $ACT$ $\gets$ create object dictionary for sample actions.
\State $REF$ $\gets$ create relations between actions by loading action network script.
\State \emph{loop:} over $P$, start with a random action from $ACT$, choose next action according to $REF$
\State \Return $P$
\EndProcedure
\end{algorithmic}
\end{algorithm}

\subsubsection{Circumplex model and Score Mapping}
\begin{figure*}[ht]
    \centering
    \includegraphics[width=0.6\linewidth]{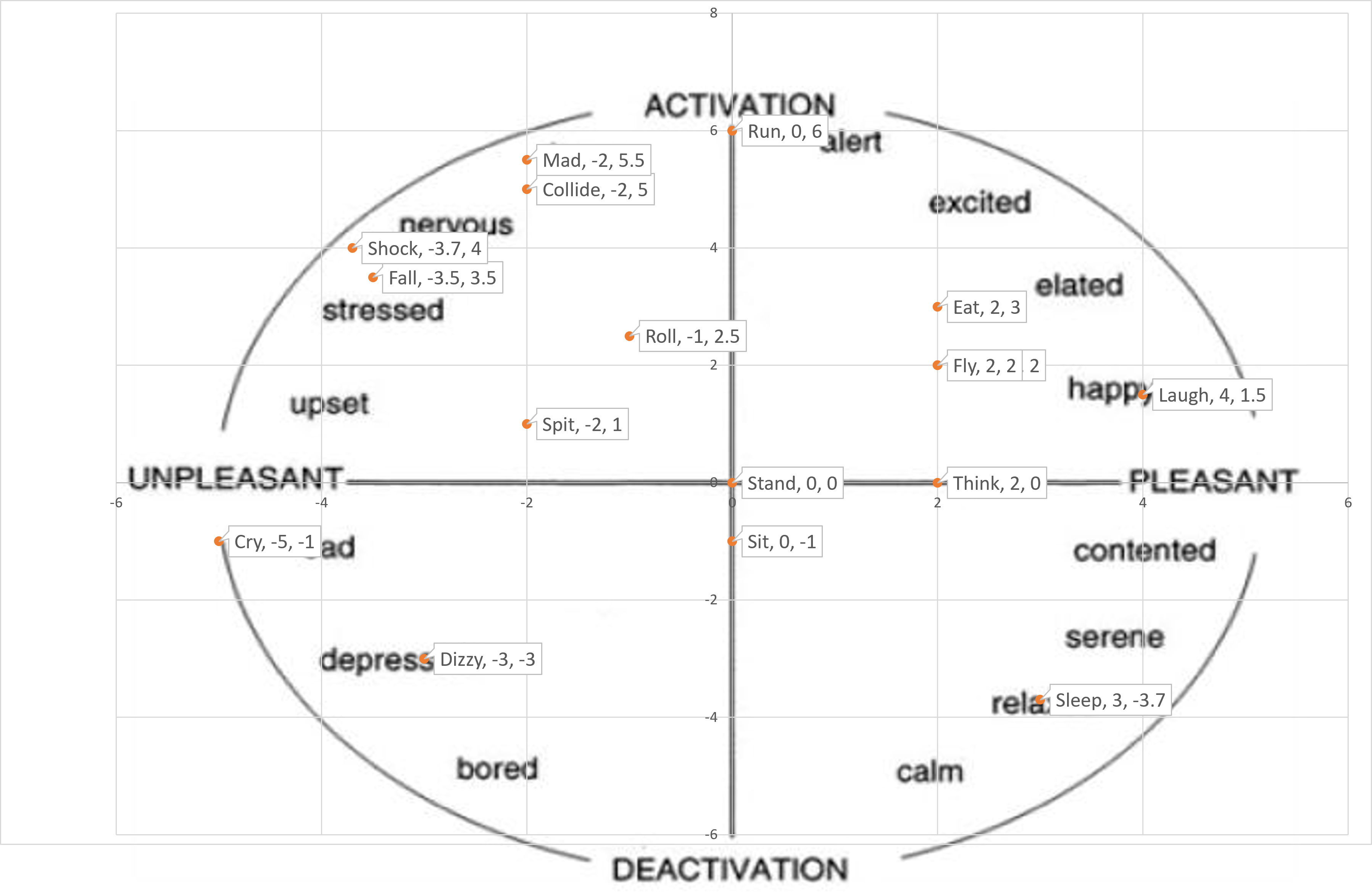}
    \caption{ Circumplex model of affect, and the actions mapping according to their related emotion.}
    \label{affect}
\end{figure*}

Our model maps the actions with narrative structure and arc score to bridge the actions with narrative arc scores. We considered the possible emotion related to the actions and then borrowed the circumplex model of affect used in emotion classification to quantify the actions. In a circumplex model of affect \cite{posner2005circumplex}, the horizontal axis representing the valence dimension, and the vertical axis representing the arousal or activation dimension, we borrow the concept to imply how much tension, if it is denoted as a number,  an action could change. Figure \ref{affect} shows how we linked the actions with emotion. For example, "laughing" is an expression of "happy," so their coordinates are arranged in the same position. Similarly,  "relief" and "sleep" are close to "relax" because they are common reactions when people feel relaxed. In contrast, action like "stand" is relatively neutral hence to be set to (0,0).

Then, we used the actions to adjust the current state to approach the narrative arc. Table\ref{modification} shows an example: a comic sequence with narrative structure [E, L, P, R]; if we map the structure into scores on a scale of 1 to 10 and use it to decide the narrative arc, it will be [2, 5, 8, 3] $+ (-1)^k \times l$, where $l$ denotes a small modification range to create some oscillations on the curve that formed by tension score; and the $k$ is a random number. The Establish(E) and the Release(R) mapped to the Exposition and Falling Action correspondingly in Freytag's Pyramid structure \cite{freytag1908freytag} of narrative arc, so their base score is relatively lower. On the contrary,   the Prolongation(L) and Peak(P) were linked to the rising action and climate, hence higher scores.

In table\ref{modification}, starting from the first panel, which scored $2 + modification$, the character performs "stand" action. Among the possible action list, because the score in the next panel is $5 + modification$, which implies the tension should go up. The extent should be close to the score difference of the two panels. "run" and "jump" will have a higher chance of being chosen because they increase the tension score, which is closer to the wanted trend of the narrative arc. We chose actions based on possibility instead of selecting by the score only because we wanted to keep the chance that the model could expand the different stories' content. The probability that an action will be chosen is 
$ \frac{1}{distance(\frac{|s_{i+1} - s_{i}|^2}{s_{i+1} - s_{i}}, a_{j})} / \sum_{\frac{1}{distance}} $.The $s_{i}$ represents the expected score of panel $i$, and $a_{j}$ denotes the activation value of action $j$. And the distance between the wanted score and action value decides the probability. The s represent  The figures in table \ref{modification} also show the expected score trend and the result of selected actions. We can see that the action selection of each character followed the curve trend.

\begin{table}
\centering
\begin{tabular}{|p{1cm}|p{1.5cm}|p{1.5cm}|p{1.5cm}|p{1.5cm}|p{2.5cm}|}
    \hline
    Set\# & E & L& P& R & narra-arc\\
    \hline
    1
    &\includegraphics[width = \linewidth]{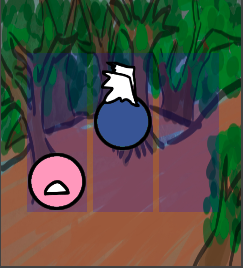}
    &\includegraphics[width = \linewidth]{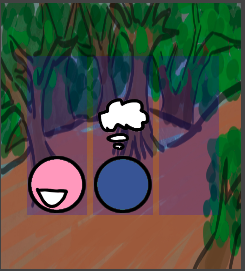}
    &\includegraphics[width = \linewidth]{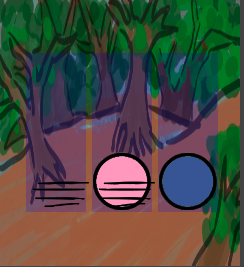}
    &\includegraphics[width = \linewidth]{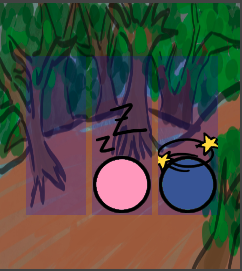}
    &\includegraphics[width = \linewidth]{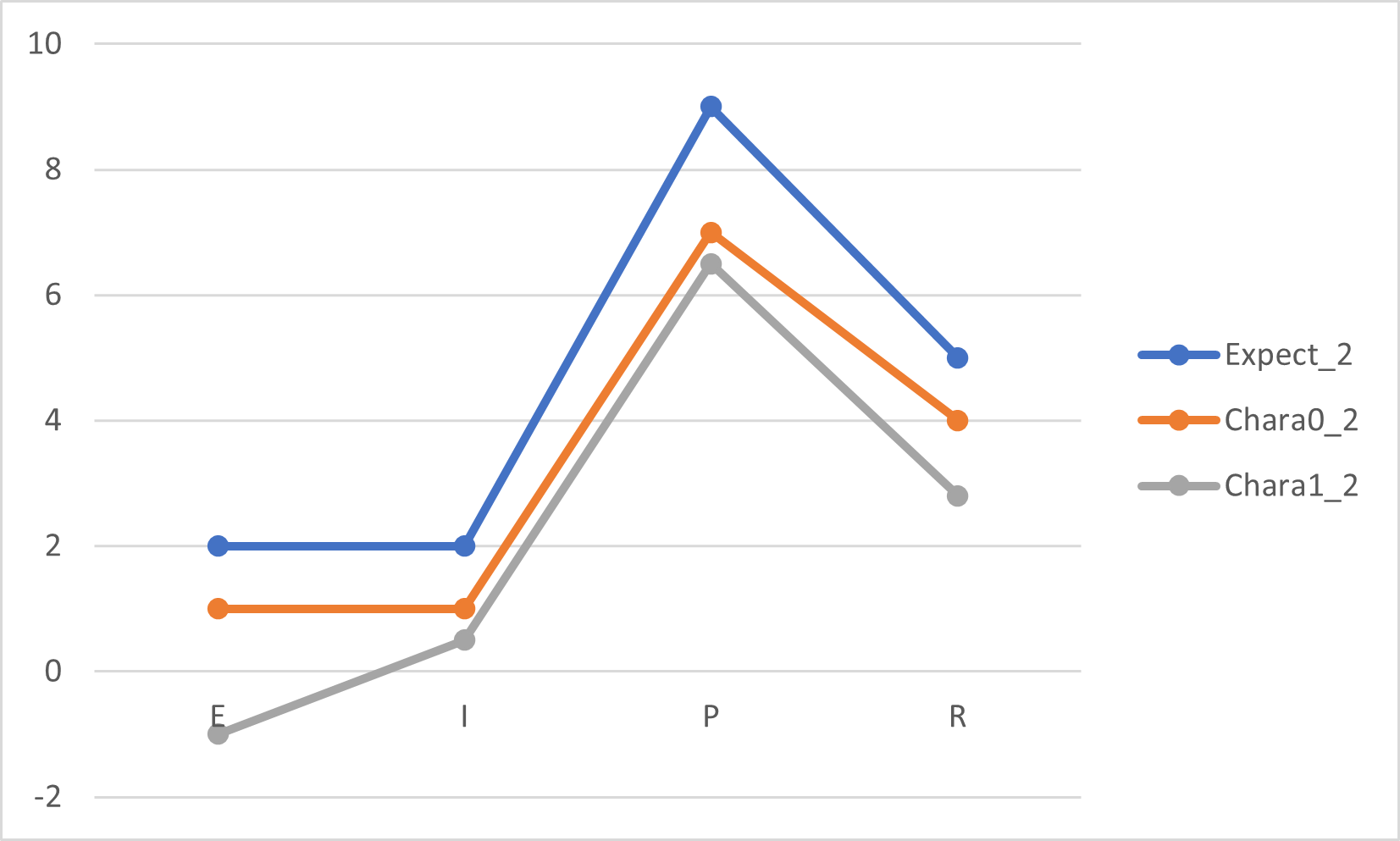}\\
    \hline     
    2
    &\includegraphics[width = \linewidth]{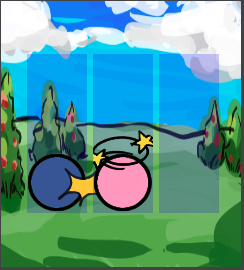}
    &\includegraphics[width = \linewidth]{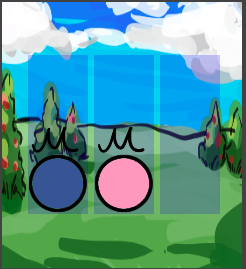}
    &\includegraphics[width = \linewidth]{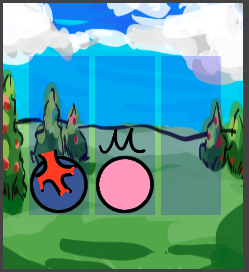}
    &\includegraphics[width = \linewidth]{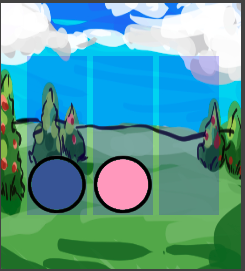}
    &\includegraphics[width = \linewidth]{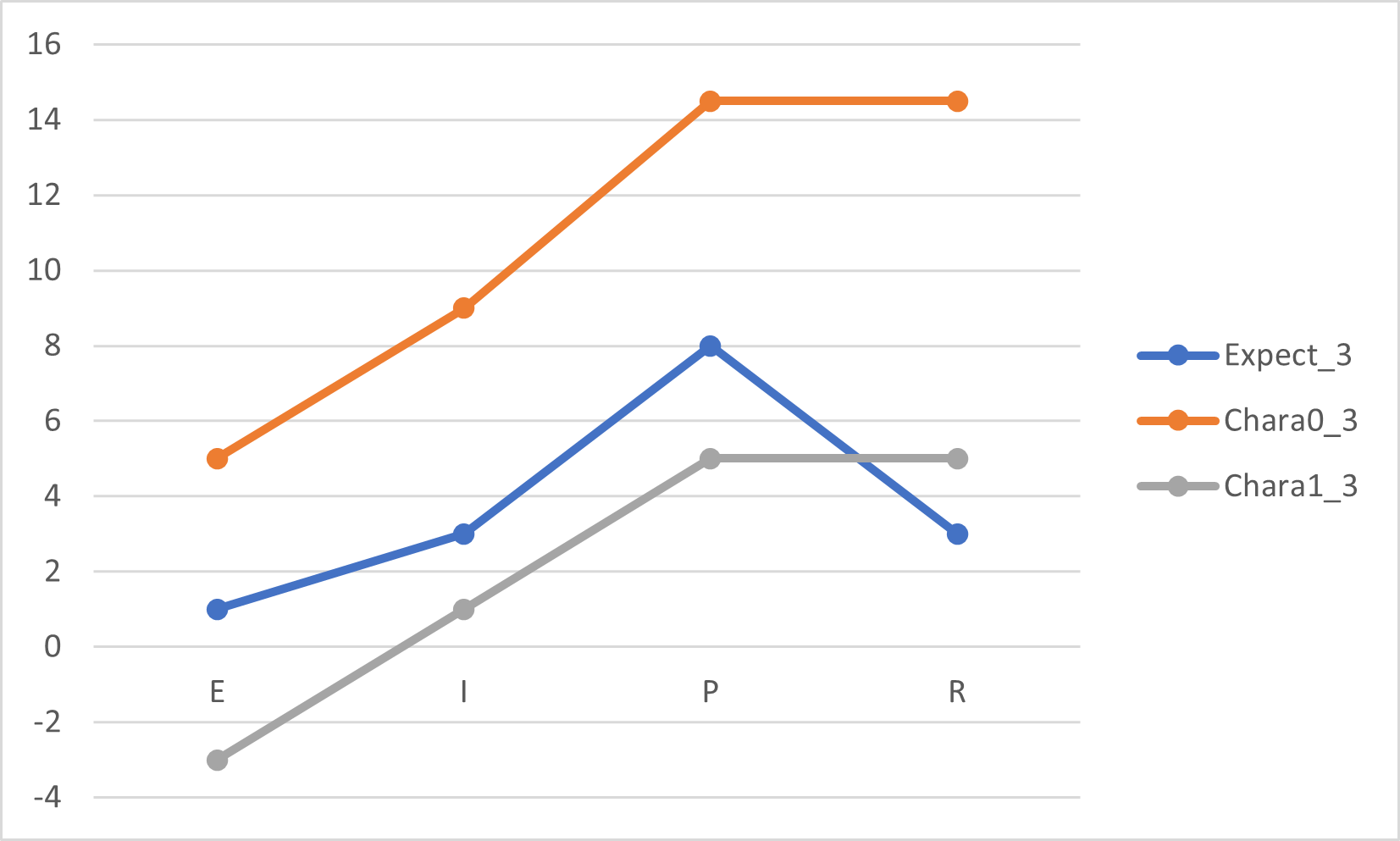}\\
    \hline     

\end{tabular}
    \caption{
    \label{modification} \small{Examples of applying action modification according to the narrative arc; and the action selection results follow tension score changes.}
    }
\end{table}

\textbf{Sequence Scoring Layer Customization}

By mapping emotion with actions, this layer combines the narrative and content by referencing the sequence's narrative arc from the previous layer \ref{score_mapping}. Inspired by the  Circumplex model, we project each action on the circle and combine it with values. After that, using the curve from the narrative arc, the layer computes the changing on the curve and selects possible actions that can fit the change most. It also sets a likelihood of taking tolerance for expanding potential action pools.

\begin{algorithm}
\caption{Narrative Arc Mapping}\label{score_mapping}
\begin{algorithmic}[1]
\Procedure{Generate Transition Sequence}{}
\State $P$ $\gets$ input the current panel sequence.
\State $ACT$ $\gets$ create value dictionary for sample actions, according to Circumplex model.
\State $NET$ $\gets$ get the action network.
\State $ARC$ $\gets$ get referenced Narrative Arc.
\State \emph{loop:} over $P$, check next panel and compute the score difference, according to $ARC$
\State Select actions in likelihood from $ACT$ and $NET$
\State \emph{loop:} revise action selection other panels according to $NET$
\State \Return $P$
\EndProcedure

\end{algorithmic}
\end{algorithm}

\subsubsection{Rendering Layer}
To enrich the graphical aspects of the comic sequence, we construct a sample layer with other specialized elements of the comic--the text balloons, as a refinement layer.

\textbf{Rendering Layer Customization}
We added customized materials to construct the layer \ref{customized_content}. Text balloons are also an important part of comic content because comics are a multi-modality medium. This layer is not complicated; it sets the possible position for text balloons and then turns on or off the display flag randomly. 

\begin{algorithm}
\caption{Customize}\label{customized_content}
\begin{algorithmic}[1]
\Procedure{Generate Transition Sequence}{}
\State $P$ $\gets$ input the current panel sequence.
\State $G$ $\gets$ create object dictionary of new materials.
\State $R$ $\gets$ set rules to assign new materials.
\State (for our textbox case) set new material positions to characters' top. 
\State \emph{loop:} select an textbox from $G$ and decide whether to display according to certain probability.
\State \Return $P$
\EndProcedure

\end{algorithmic}
\end{algorithm}

\subsubsection{Story Content}
To apply customized story content in the generating process, we implemented a sample layer that uses customized content as a reference to generate panel content. \ref{story_content} gives the steps of how this layer is designed.

\textbf{Story Content Layer Customization}
To process the story text, we incorporate a language processing framework--Rensa\cite{harmon2017narrative}, which encodes multiple narrative representations in a human-readable format. It extracts story text as "assertions," triples that consist of left (subject), right (object), predicates, tense, and properties. The predicates are the relations between the left and right. For example, an assertion from the text "Alice and her sister is in the forest." becomes $\{"l": ["Alice," "Sister"], "relation": ["location_on"], "r": ["forest"] \}$. Therefore, in our script for this story content layer, the process started with loading and sorting the assertions. The model retrieves the ones with character and location information as the reference for comic subjects and the scene. Then, it modifies the panel content according to the story text's actions.

\begin{algorithm}
\caption{Customize}\label{story_content}
\begin{algorithmic}[1]
\Procedure{Modify Content According to Story}{}
\State $A$ $\gets$ import assertions of story text from the tool Rensa.
\State sort the $A$.
\State Add links with story content (character, actions, and scene).
\State \emph{loop:} modify panel content according to the links.
\State \Return $P$
\EndProcedure

\end{algorithmic}
\end{algorithm}

\section{Generated Example with Customization}

We designed structure refinement layers through VNG and mapped the categories to curves that express narrative-arc to improve action selection for creating content. And another refinement layer employed comic transitions to increase visual variety on panels. Moreover, we applied two extension layers to add richness to image content, Which led to showcases for sample layers.

We first compared generated sequences based on a narrative arc decided by VGN or sequences generated freely. This showcases aims to compare the content differences caused by following the narrative arc. The second part displayed the changes led by the underlying rules applied when constructing content. In other words, this aims to show the possibility of altering the constructing layer to influence content details. Besides the overall structure and component of panels, we also wanted to test the comic sequence's potential by modifying the links between panels. The third part showed the flexibility in controlling the discourse affected by the inter-panel adjustment. The fourth part showed the possibility of extending the generator to enhance the richness of content expression.

\subsection{Grammar Layer and Action Selection Based on Narrative Arcs}

Our first two refinement layers implement the five categories of VGN and expand the center-embedded structure through probability for the whole comic panel. Then, according to the usages that the categories suggested, we mapped the categories to stages in narrative arcs. Then the action candidates that in our dataset utilized the activation and deactivation concept to get a corresponding coordinate in the Circumplex model, figure \ref{affect}. The coordinate implies whether an action raises tensions or drops tensions.


Therefore, the expanded narrative grammar tree basically decides how the story looks in the generated comic sequence. The table\ref{grammar_layer_and_arc} shows the results of whether the grammar and narrative arc layer is on or off.
		
\begin{table}
\centering
\begin{tabular}{|p{1.5cm}|p{6cm}|p{2.5cm}|}
    \hline
    Settings & Layer: Narrative Grammar and Narrative Arc & Narrative Trend\\
    \hline
   with
    &\includegraphics[width = \linewidth]{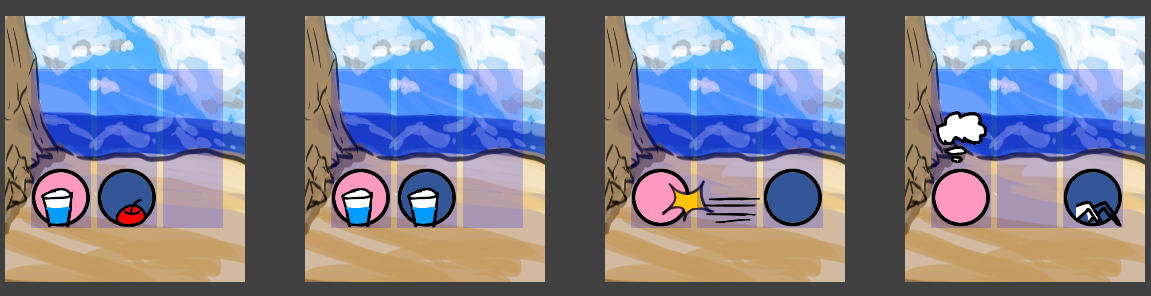}
    &\includegraphics[width = \linewidth]{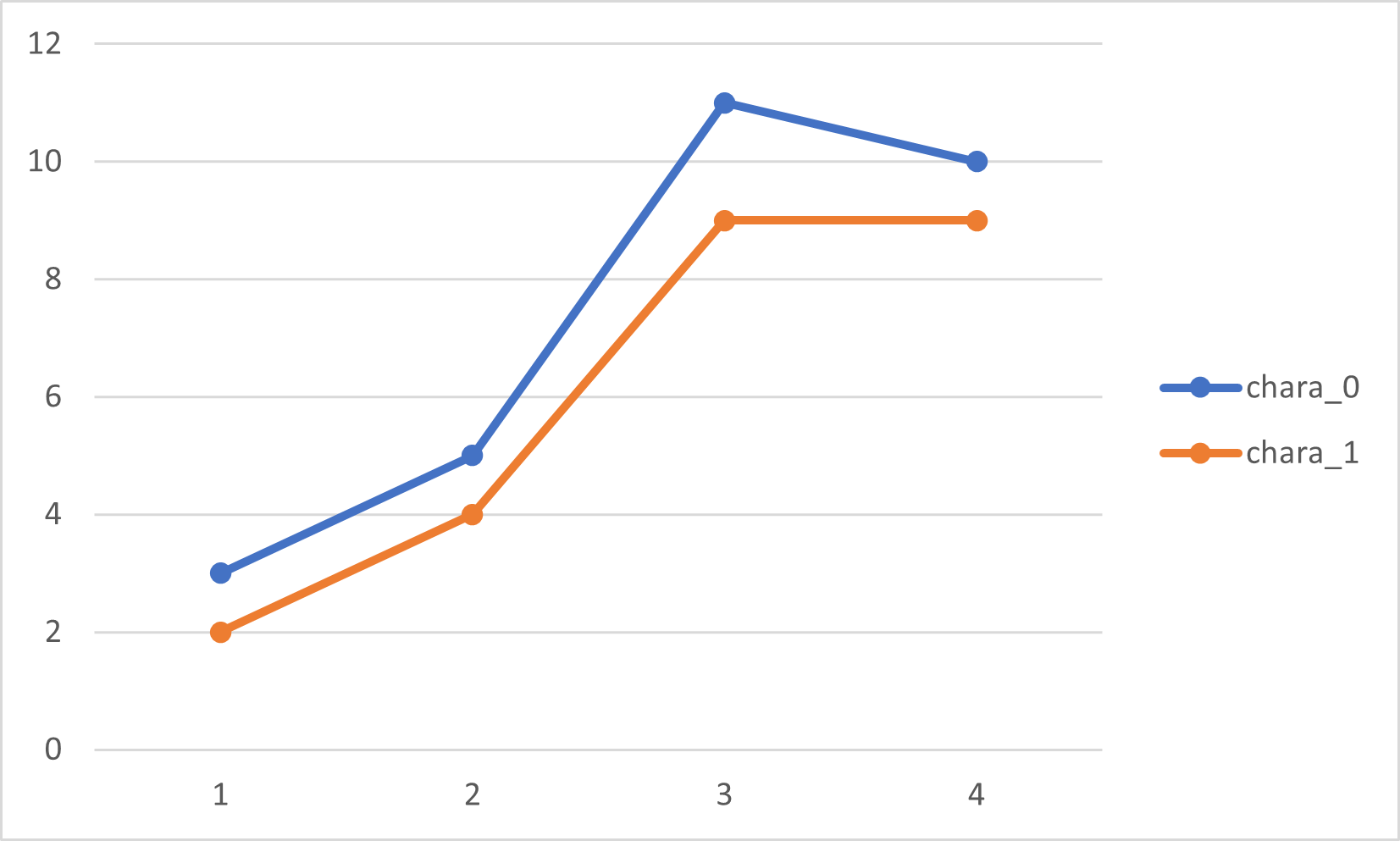}\\
    \hline     
   without
    &\includegraphics[width = \linewidth]{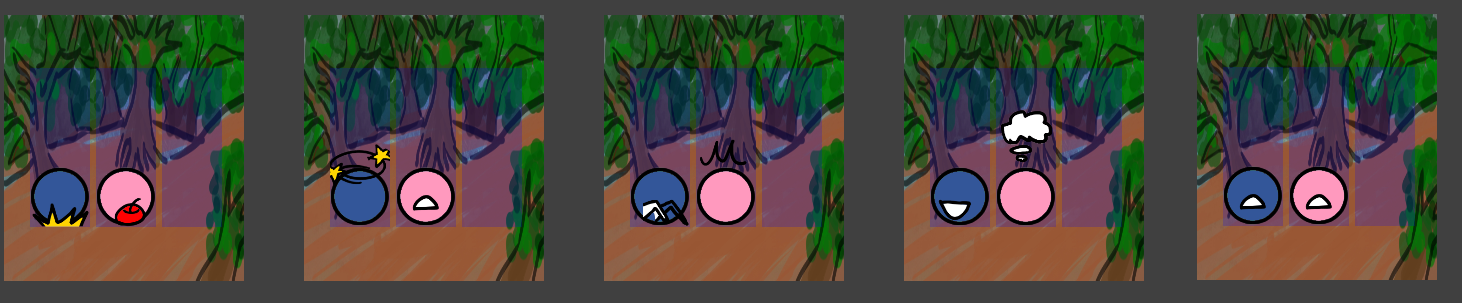}
    &\includegraphics[width = \linewidth]{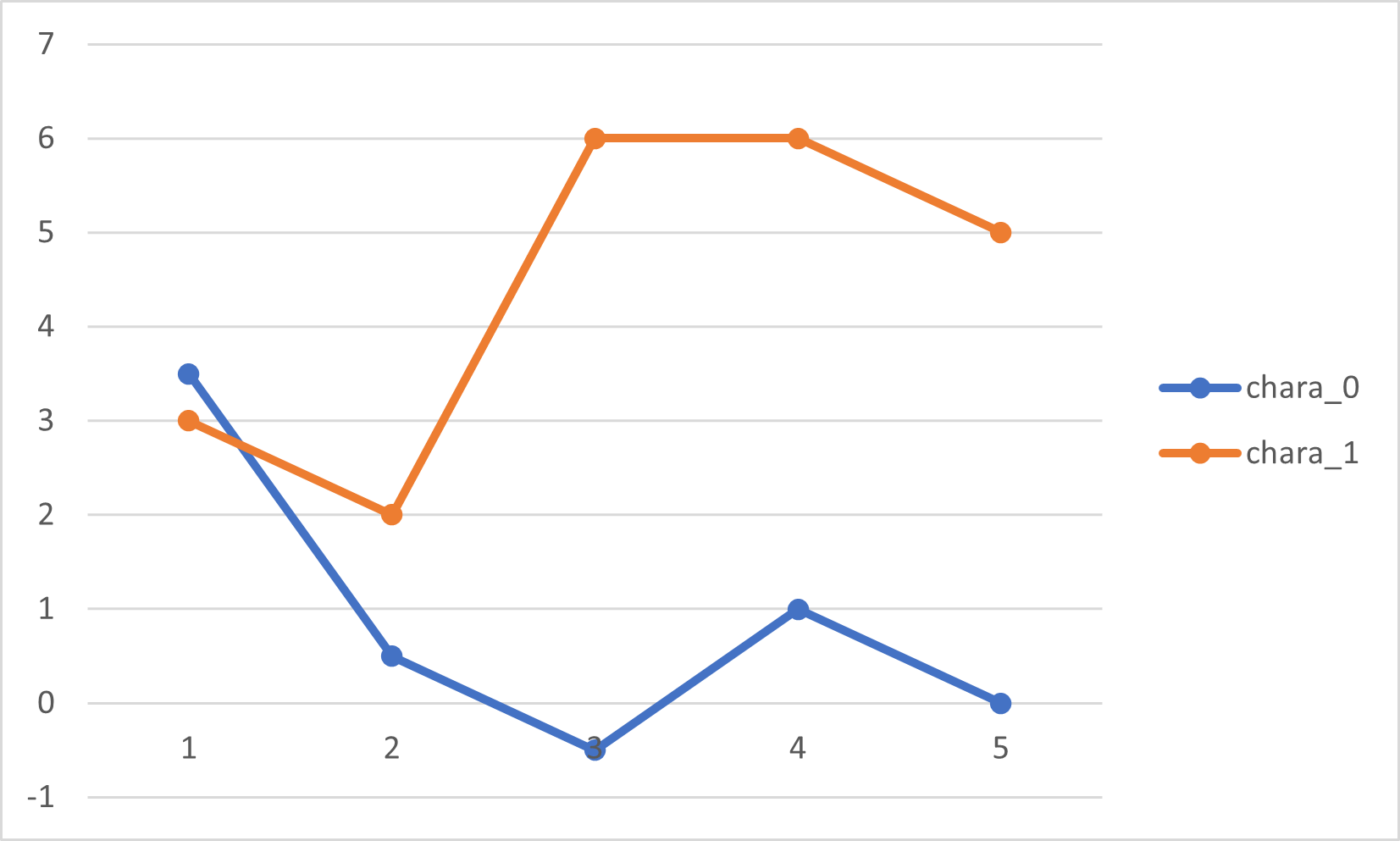}\\
    \hline     

\end{tabular}
    \caption{
    \label{grammar_layer_and_arc} \small{Examples of with or without narrative grammar and narrative structures.}
    }
\end{table}
The sequence with the narrative-arc layer suggests that the action selection follows a usual pattern. The action has higher tension such as collide and run showed up in peak and then tension relieved by more relaxed actions. And both the action of the characters follow the trend of the tension curve. On the contrary, the characters' action selection becomes rather arbitrary in the sequence without refining the narrative-arc layer. Even in the story peak, the characters take relaxed actions.

\subsection{Action relations network}
The next refinement layer implements are action-reaction relation network of possible actions. For example, the "fall" happens after "jump" or "fly," "dizzy" happens after "collide," and so on. The table \ref{action_net_layer} suggests the results where the action relation network was on or off.
\begin{table}
\centering
\begin{tabular}{|p{1.5cm}|p{6cm}|}
    \hline
    Settings & Layer: Action Relations Network Layer \\
    \hline
   with
    &\includegraphics[width = \linewidth]{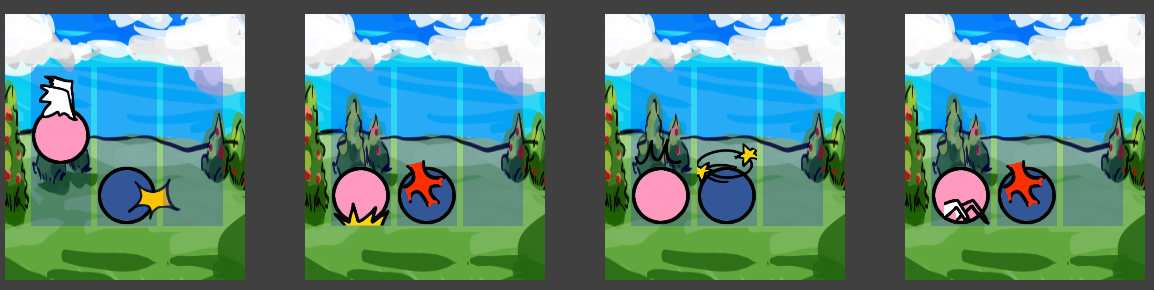}\\
    \hline     
   without
    &\includegraphics[width = \linewidth]{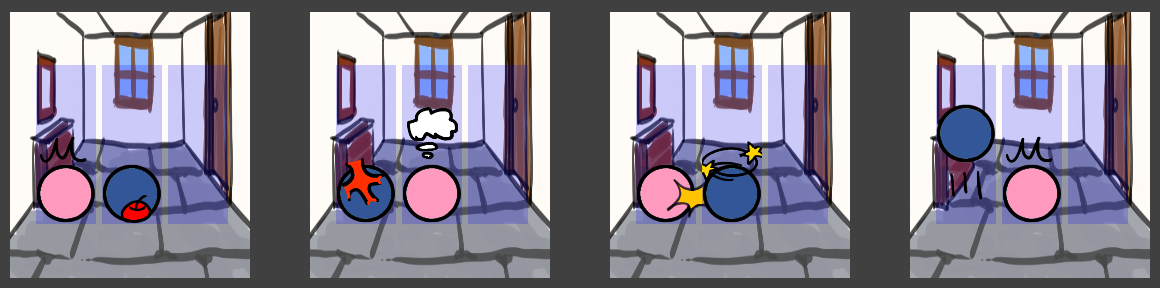}\\
    \hline     

\end{tabular}
    \caption{
    \label{action_net_layer} \small{Examples of with or without action relation network.}
    }
\end{table}
The sequence with the action relation network can easily link the panel with a continuous story, whereas the other can not. In the former, the blue character "collide" something, so it becomes "angry" and then "dizzy The pink character "fly" and then "falls." After that, it is "shocked" and then takes "a rest." On the contrary, characters in the later sequence select actions randomly.

\subsection{Transition layer}
The transition refinement layer uses the transitions to modify the story presentation of the comic sequence. For example, in the table \ref{transition_layer}, the sequence linked by [Aspect, Action, Scene] transitions, adjusted the image's composition, modifies the chosen actions, and transports a significant space. 
The view angle changes make the story a bit more interesting. The modification makes the story seem to be interpreted as two characters meeting in a garden, and one suddenly leaves the place. When the other follows the character, it saw the character sit in the forest. 
\begin{table}
\centering
\begin{tabular}{|p{1.5cm}|p{6cm}|}
    \hline
    Settings & Layer: Transition layer \\
    \hline
   with
    &\includegraphics[width = \linewidth]{Images/transition_yes.PNG}\\
    \hline
   transition sequence
    &[Aspect-to-Aspect, Action-to-Action, Scene-to-Scene]\\
    \hline      
   without
    &\includegraphics[width = \linewidth]{Images/action_yes.PNG}\\
    \hline     

\end{tabular}
    \caption{
    \label{transition_layer} \small{Examples of with or without transition network.}
    }
\end{table}
\subsection{Customize layers}
The final two refinement layers are the cases that extend the generator with either image set or functions. The textbox layer provides three types of text balloon where the sharpness degree of ballon's edge help to emphasize the emotion of characters. The action with higher activation scores will be pairing the sharp ballon when the mild action will be paired with the normal talking ballon. Furthermore, the shapes present whether the character is talking or thinking. The display layer implemented the function that flips the display flag of characters on or off by probability. In the last two panels, as the blue character's display flag was turned off, the whole sequence then shows a story that two characters flying together, but one disappeared eventually. Both of the extended layers enrich the image content and add more detail to the presented story. The results are in table \ref{addtional_layer}  
\begin{table}
\centering
\begin{tabular}{|p{1.5cm}|p{6cm}|}
    \hline
    Settings &  \\
    \hline
   Textbox layer
    &\includegraphics[width= \linewidth]{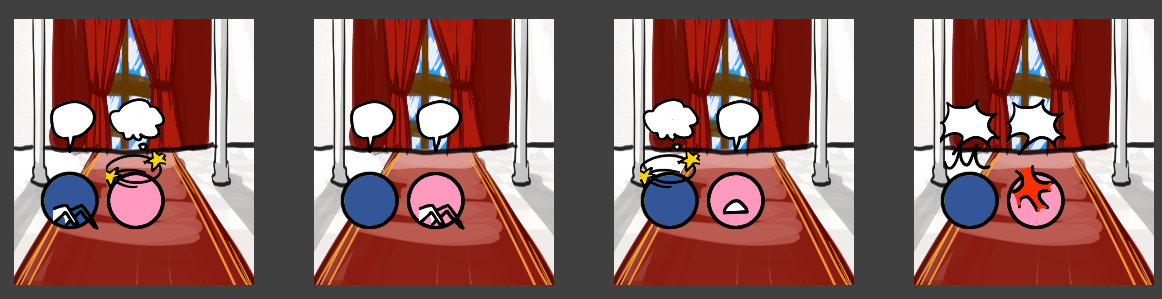}\\
    \hline     
    
   Display layer
    &\includegraphics[width = \linewidth]{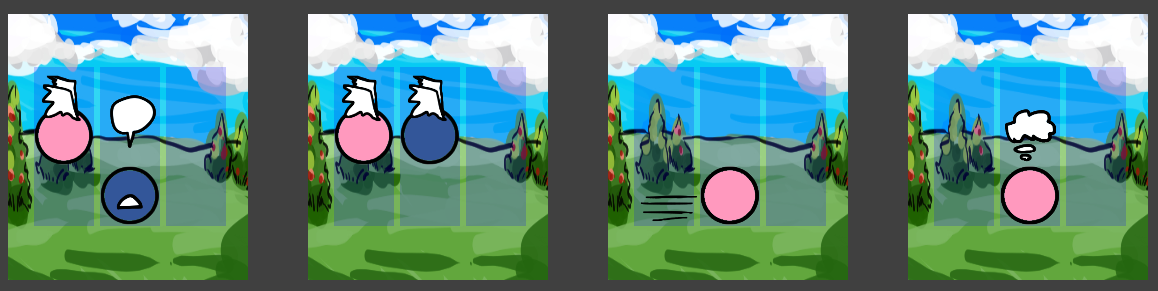}\\
    \hline     

\end{tabular}
    \caption{
    \label{addtional_layer} \small{Examples that apply additional refinement layers }
    }
\end{table}

\subsection{Story layers}
We used short story text as a content reference in our final customized layer. Our model didn't equip language processing functions, so this layer starts from "assertions": the proceed triples from the narrative representation model--Rensa. The sample story text is below. and the assertions are in table \ref{assertions}.

\{A and B are in the forest. A has B's Apple. B is angry. B enter the scene.\}

As the table \ref{assertions} shows, some links between character, relation, and location are created. And then, the generator modifies the content according to the links. It selects performable actions from the linked action pool. For example, the relation "has" is related to actions like eating and drinking; and the action "angry" links to other action labels such as crying and dizziness. The results are in table \ref{story_layer}.

\begin{table}
\centering
\begin{tabular}{|p{1.5cm}|p{3cm}|p{1.5cm}|p{1.5cm}|p{4cm}|}
    \hline
    l &  relation & r & r\_owner & links\\
    \hline
    \{A, B\} & \{location\_on\} & \{forest\} & \{\} & \{forest\} \\
    \hline
    \{A\} & \{has\} & \{apple\} & \{B\} & \{eat, drink\} \\
    \hline
    \{B\} & \{action\} & \{angry\} &\{\} & \{angry, dizzy, cry\} \\
    \hline
    \{B\} & \{action\} & \{enter\} &\{\} & \{run, collide\} \\
    \hline     

\end{tabular}
    \caption{
    \label{assertions} \small{Examples that apply short story text as reference.}
    }
\end{table}

The example gives a complete story that we can interpret as follows: the two characters are in the forest, and are both eating. Then after one of them finished its apple, it took the other's apple. So, the other character became angry. Then the first character ran, and the other character chased it. Because the content was created depending on the links between story text and links between existing actions in the dataset, it will follows the story flow.

\begin{table}
\centering
\begin{tabular}{|p{1.5cm}|p{6cm}|}
    \hline
    Settings &  \\
    \hline
   Story layer
    &\includegraphics[width= \linewidth]{Images/story_example.png}\\
    \hline     

\end{tabular}
    \caption{
    \label{story_layer} \small{Examples that apply additional refinement layers }
    }
\end{table}

\section{Discussion}
In the working example, we demonstrated the variety of comics that are generated based on idioms encoding comic theories. The VNG is useful for the progression of comic content that follows certain patterns. The use of action relations enhanced the coherence of the comic sequence.  The transitions between panels controlled the pace of content as well as scene or character state transitions. The composition templates and the collected symbolic metaphor increases the dynamics of panel images. 

There are many possibilities for combining the refinement layers for potential authors of comics using this toolset. For the overall progression pattern, because our narrative arc depended on the center-embedded structures generated from VNG, the possibility of structure can be presented as $2^4 \times (2^5 \times 2^4)^{n-1}$. Given that except the Peak(P), all other categories are optional, this leads to the 4 power of 2 possibilities. Each category can potentially expand further with grammar sequence to get a new layer, thus causes each new layer to have 5 power of 2's possible sequences. 
 
With our sample action relation network, suppose we have $p$ actions and represent the network as a directed graph;  the upper bound of the possible action sequence that constructs the content will be $\frac {(p - 2)\,!}{(p - 2 - k)\,!}$ if the sequence length is $k$. The reasonable links between action and their possible reactions will graph a subset of the complete graph with p nodes. Therefore, the actions can only create at most the number of possible paths between two verticles with length $k$.  In our sample network, the possibility will be $\frac {{17\,!}}{{12\,!}}$


\section{Conclusion}
This paper demonstrates a tool that encodes idioms from comic theories for generation. We showed the variety of sequences that can be generated in terms of panel diversity. Even with a small hand-crafted ontology, we show that the sequential process of authoring across layers is effective in generating diversity.  

Our contributions include an initial content set-- verbs, symbolic metaphors, characters, scenes, composition templates, action-icon combination, and action relations. There are several avenues of potential future works. For example, combining the generator with intelligent interfaces can assist authors in creating comics by giving suggestions based on comic theories as a mixed-initiative interactive authoring tool. The modularity of the process allows research modules to be independently incorporated. For example, a plot generator could be separately plugged in to a panel composition algorithms.

\bibliographystyle{natdin}
\bibliography{Rimi_reference}

\end{document}